\documentclass[sigconf]{acmart}

\usepackage{multirow}

\usepackage{subfigure}

\usepackage{color, colortbl}
\definecolor{Gray}{gray}{0.9}
\definecolor{LightCyan}{rgb}{0.88,1,1}

\usepackage[normalem]{ulem}

\AtBeginDocument{%
  \providecommand\BibTeX{{%
    \normalfont B\kern-0.5em{\scshape i\kern-0.25em b}\kern-0.8em\TeX}}}

\copyrightyear{2023}
\acmYear{2023}
\setcopyright{acmlicensed}\acmConference[WWW '23]{Proceedings of the ACM Web Conference 2023}{May 1--5, 2023}{Austin, TX, USA}
\acmBooktitle{Proceedings of the ACM Web Conference 2023 (WWW '23), May 1--5, 2023, Austin, TX, USA}
\acmPrice{15.00}
\acmDOI{10.1145/3543507.3583294}
\acmISBN{978-1-4503-9416-1/23/04}

\begin{document}

\title{LED: Lexicon-Enlightened Dense Retriever for \\ Large-Scale Retrieval}

\settopmatter{authorsperrow=4}
\author{Kai Zhang}
\authornote{Work done during the internship at Microsoft.}
\affiliation{%
  \institution{The Ohio State University}
  \city{Columbus}
  \state{Ohio}
  \country{USA}
}
\email{zhang.13253@osu.edu}

\author{Chongyang Tao}
\affiliation{%
  \institution{Microsoft Corporation}
  \city{Beijing}
  \country{China}
}
\email{chotao@microsoft.com}

\author{Tao Shen}
\affiliation{%
  \institution{AAII, FEIT, University of Technology Sydney}
  \city{Sydney}
  \country{Australia}
}
\email{tao.shen@uts.edu.au}

\author{Can Xu}
\affiliation{%
  \institution{Microsoft Corporation}
  \city{Beijing}
  \country{China}
}
\email{caxu@microsoft.com}

\author{Xiubo Geng}
\affiliation{%
  \institution{Microsoft Corporation}
  \city{Beijing}
  \country{China}
}
\email{xigeng@microsoft.com}

\author{Binxing Jiao}
\affiliation{%
  \institution{Microsoft Corporation}
  \city{Beijing}
  \country{China}
}
\email{binxjia@microsoft.com}

\author{Daxin Jiang}
\authornote{Corresponding author.}
\affiliation{%
  \institution{Microsoft Corporation}
  \city{Beijing}
  \country{China}
}
\email{djiang@microsoft.com}




\begin{abstract}
Retrieval models based on dense representations in semantic space have become an indispensable branch for first-stage retrieval. 
These retrievers benefit from surging advances in representation learning towards compressive global sequence-level embeddings. However, they are prone to overlook local salient phrases and entity mentions in texts, which usually play pivot roles in first-stage retrieval.
To mitigate this weakness, we propose to make a dense retriever align a well-performing lexicon-aware representation model. The alignment is achieved by weakened knowledge distillations to enlighten the retriever via two aspects -- 1) a lexicon-augmented contrastive objective to challenge the dense encoder and 2) a pair-wise rank-consistent regularization to make the dense model's behavior incline to the other. We evaluate our model on three public benchmarks, which shows that with a comparable lexicon-aware retriever as the teacher, our proposed dense one can bring consistent and significant improvements, and even outdo its teacher.  In addition, we show our lexicon-aware distillation strategies are compatible with the standard ranker distillation, which can further lift state-of-the-art performance.\footnote{Code is available at https://github.com/drogozhang/LED.}
\end{abstract}


\maketitle
\title{LED: Lexicon-Enlightened Dense Retriever for Large-Scale Retrieval}

\section{Introduction}
Large-scale passage retrieval~\cite{Cai2021IRSurvey} aims to fetch relevant passages from a million- or billion-scale collection for a given query to meet users' information needs, serving as an important role in many downstream applications including open domain question answering~\cite{Karpukhin2020DPR}, search engine~\cite{Zou2021Baidu}, and recommendation system~\cite{Zhang2019RSSurvey}, etc.
Recent years have witnessed an upsurge of interest and remarkable performance of dense passage retrievers on first-stage retrieval. Built upon powerful pre-trained language models (PLM)~\cite{Devlin2019BERT, Liu2019RoBERTa, Radford2018GPT}, dense retrievers~\cite{Karpukhin2020DPR, Xiong2021ANCE,Qu2021RocketQA} encode queries and passages into a joint low-dimensional semantic space in a Siamese manner (i.e. dual-encoder), so that the passages could be offline pre-indexed and query could be encoded online and searched via approximate nearest neighbor~\cite{John2021ANN}, reaching an efficiency-effectiveness trade-off.

Although dense retrieval becomes indispensable in modern systems, a long-term challenge is that the dense representations in a latent semantic space are abstractive and condensed, exposing the systems to a risk that pivot phrases and mentions may be overlooked and thus leading to sub-optimal efficacy. For example, DPR~\cite{Karpukhin2020DPR} didn't regard ``\textit{Thoros of Myr}'' as an entity mention in the query ``\textit{Who plays Thoros of Myr in Game of Thrones?}''. Analogously, given the query ``\textit{What is an active margin}'', ANCE~\cite{Xiong2021ANCE} overlooked the ``\textit{active margin}'' as an entire local salient phrase and hence retrieved passages related to the financial term ``\textit{margin}''. As a remedy, prior works resort to either coupling a dense retriever with the term matching scores (e.g., TF-IDF, BM25) \cite{Gao2021COIL, Lin2021UniCOIL, Dai2019DeepCT, Mallia2021DeepImpact} or learning BM25 ranking into a dense model as additional features to complement the original one~\cite{Chen2021ImitateSparse}. But, these approaches are limited by superficial combinations and almost unlearnable BM25 scoring.

To circumvent demerits from the superficial hybrid or learning with inferior lexicon-based representations upon PLM, we propose a brand-new lexicon-enlightened dense (LED) retriever learning framework to inject rich lexicon information into a single dense encoder, while keeping its sequence-level semantic representation capability. 
Instead of prevailing BM25 as lexicon-rich sources, we propose to leverage the recently advanced lexicon-centric representation learning model transferred from large-scale masked language modeling (MLM), and attempt to align a dense encoder with two brand-new weakened distilling objectives. On the one hand, we present lexicon-augmented contrastive learning that incorporates the hard negatives provided by lexicon-aware retrievers for contrastive training. Intuitively, the negatives given by the lexicon-aware models could be regarded as adversarial examples to challenge the dense one, so as to transfer lexical knowledge to the dense model.
On the other hand, inspired by previous work~\cite{Burges2005RankNet, Dehghani2017WeakSupervision}, we propose a pair-wise rank-consistent regularization as a weak supervision to guide dense model’s behavior incline to the lexicon-aware ones. Compared to distribution regularization such as KL-divergence~\cite{Zhang2021AR2} and strict fine-grained distillation like Margin-MSE~\cite{Hofstatter2020Margin-MSE}, LED provides weak supervision signals from the lexicon-aware retrievers, leading to desirable partial knowledge injection while maintaining the dense retriever's own properties.

We evaluate our method on three real-world human-annotated benchmarks. Experimental results show that our methods consistently and significantly improve dense retriever's performance, even outdoing its teacher.
Notably, these significant improvements are brought by the supervision of a performance-comparable lexicon-aware retriever. 
Besides, a detailed analysis of retrieval results shows that our knowledge distillation strategies indeed equip the dense retriever with lexicon-aware capabilities.
Lastly, we show our lexicon-aware distillation strategies are compatible with the standard ranker distillation, achieving further improvement and a new state-of-the-art performance.

Our contributions are three-fold: 
(1) We consider improving the dense retriever by imitating the retriever based on the lexicon-aware representation model upon PLM; 
(2) We propose two strategies including lexicon-augmented contrastive training and pair-wise rank-consistent regularization to inject lexical knowledge into the dense retriever; 
(3) Evaluation results on three benchmarks show that our method brings consistent and significant improvements to the dense retriever with a comparable lexicon-aware retriever as a teacher and a new state-of-the-art performance is achieved.

\section{Related Work}

\begin{figure*}[t]
    \centering
    \includegraphics[width=0.92\textwidth]{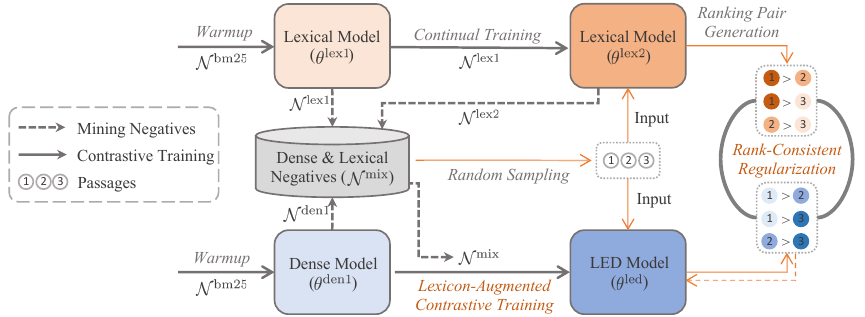}
    \caption{The training framework of our LED retriever. The Lexical teacher is independently trained following a two-stage process.
    After warming up, LED is trained with negatives mined by self and two lexicon-aware retrievers for lexicon-augmented contrastive learning, during which the Lexical Model ($\theta^{\mathrm{lex2}}$) enhances LED with pair-wise rank-consistent regularization.}
    \label{fig:training-process}
\end{figure*}

Current passage retrieval systems are widely deployed as retrieve-then-rank pipelines~\cite{Huang2020Embedding, Zou2021Baidu}. The first-stage retriever (i.e., dual-encoder)~\cite{Xiong2021ANCE, Qu2021RocketQA, Ren2021PAIR, Wu2022CoT-MAE, Lin2022Aggretriever, Lin2022PROD} selects a small number of candidate passages (usually at most thousands) from the entire collection, and the second-stage ranker (i.e., cross-encoder~\cite{Zhou2022R2anker}) scores these candidates again to provide a more accurate passages order. In this paper, we focus on enhancing the first-stage retriever.

\paragraph{\textbf{Dense Retriever.}}
Built upon Pre-trained Language Models~\cite{Devlin2019BERT, Liu2019RoBERTa}, dense retriever~\cite{Karpukhin2020DPR, Qu2021RocketQA} is to capture the semantic meaning of an entire sequence by encoding sequential text as a continuous representation into a low-dimensional space (e.g., 768).
In this way, the dense retriever could handle vocabulary and semantic mismatch issues within the traditional term-based techniques like BM25~\cite{Robertson2009BM25}. 
To train a better dense retriever, various techniques are proposed for providing hard negatives including reusing in-batch negatives~\cite{Karpukhin2020DPR, Luan2021MEBERT, Qu2021RocketQA}, iteratively sampling~\cite{Xiong2021ANCE}, mining by a well-trained model or dynamic sampling~\cite{Zhan2021STAR-ADORE, Zhang2021AR2}, and denoising by cross-encoder~\cite{Qu2021RocketQA}.
To build retrieval-specific pre-trained language models, \citet{Lee2019ICT} proposed an unsupervised pre-training task, namely Inverse Cloze Task (ICT), ~\citet{Gao2021Condenser} decoupled model architecture during pre-training and further designed corpus-level contrastive learning~\cite{Gao2021coCondenser} for better passage representations.

\paragraph{\textbf{Lexicon-Aware Retriever.}}
Another paradigm of work~\cite{Formal2021SPLADE, Gao2021COIL, Lin2021UniCOIL} takes advantage of strong PLMs to build lexicon-aware sparse retrievers by term-importance~\cite{Dai2019DeepCT, Lin2021UniCOIL} and top coordinate terms~\cite{Formal2021SPLADE, Formal2021SPLADEv2}. 
These models have lexical properties and could be coupled with inverted indexing techniques.
Based on contextualized representation generated by PLMs~\cite{Devlin2019BERT}, ~\citet{Dai2019DeepCT} learned to estimate individual term weights, ~\citet{Mallia2021DeepImpact} further optimized the sum of query terms weights for better term interaction, COIL-series works used token-level interactions on weight vector~\cite{Gao2021COIL} or scalar~\cite{Lin2021UniCOIL} to obtain exact word matching scores, and ~\citet{Formal2021SPLADE, Formal2021SPLADEv2} trained a retriever encoding passages as vocabulary-size highly sparse embeddings.
Recently, ~\citet{Chen2021ImitateSparse} trained a PLM-based retriever from scratch with data generated by BM25. The trained lexicon-aware retriever could encode texts as low-dimensional embeddings and have identical lexical properties and comparable performance with BM25.

\paragraph{\textbf{Hybrid Retriever.}}
Arguably, dense sequence-level retrievers and lexicon-aware retrievers have distinctive pros and are complementary to each other. This fact triggered researchers to investigate how to combine their advantages, such as direct score aggregation~\cite{Kuzi2020ScoreAgg}, weighted sum~\cite{Wang2021Interpolation, Li2022Interpolate, Luan2021MEBERT, Lin2021inbatch, Lin2021UniCOIL}, multiplication~\cite{Xu2022LaPraDoR}, or concatenation~\cite{Chen2021ImitateSparse, Seo2019DENSPI, Ma2021QGenHyb} in an ensemble system.
{The above hybrid retrievers require two dense and lexical retrievers to first compute individually and then combine their results in feature-level~\cite{Chen2021ImitateSparse, Seo2019DENSPI, Ma2021QGenHyb} or relevance-score-level~\cite{Kuzi2020ScoreAgg, Wang2021Interpolation, Li2022Interpolate, Luan2021MEBERT, Lin2021inbatch, Lin2021UniCOIL, Xu2022LaPraDoR} to obtain a final result. In contrast, our method only needs one model to achieve both dense and lexicon-aware retrieval behaviors, significantly decreasing the memory footprint and inference speed meanwhile providing a more in-depth fusion of lexicon-aware and dense retrieval views.}

\paragraph{\textbf{Knowledge Distillation.}}
Cross-encoder empirically outperforms dual-encoder for it inputs query and passage as a whole, so that attention mechanism will be applied between them, leading to in-depth token-level interactions. Its superior performance motivates many works~\cite{Hofstatter2020Margin-MSE, Zhang2021AR2} to enhance dual-encoders by knowledge distillation from cross-encoder.
KL-Divergence, which minimizes the distances of distributions between teacher and student, has proven effective in many works~\cite{Zhang2021AR2, Khattab2021ColBERTv2}.
Margin-MSE~\cite{Hofstatter2020Margin-MSE} aims to minimize the difference of margins in two passages, and it's been applied in later works~\cite{Formal2021SPLADEv2, Hofstatter2021TAS-B}.
~\citet{Reddi2021RankDistil} used the teacher's top passages as positive examples to teach students point-wisely.
ListNet~\cite{Cao2007ListNet, Xiao2022Distill-VQ} ensured the consistency of list-wise ranking order by minimizing the difference in score distributions over passages.

The above methods are designed for the cross-encoder teacher that is much stronger than the dense student. But in our situation, the lexicon-aware teacher can only achieve comparable performance with a dense retriever.
In practice (i.e., Tab.~\ref{tab:distillation}), we find that existing distillation methods may not be perfect choices in our setting. Therefore, we propose two novel strategies, namely lexicon-augmented contrastive training and pair-wise rank-consistent regularization to transfer lexical knowledge.

\section{Methodology}
We first introduce the task formalization, general training framework, and retriever architectures in Sec.~\ref{sec:preliminary}.
Then we present our \textbf{L}exical- \textbf{E}nlightened \textbf{D}ense (\textbf{LED}) retriever in Sec.~\ref{sec:LED}.

\subsection{Preliminary}
\label{sec:preliminary}
\paragraph{\textbf{Task Definition.}}
In the first-stage retrieval, given a query $q$, a retriever is required to fetch top-$k$ relevant passages from a million- even billion-scale passage collection $\mathcal{C}$.
Due to the efficiency requirement, dual-encoder architecture is widely applied in this task for its lightweight metric calculation. 
Formally, dual-encoder represents text $x$ (could be query $q$ or passage $p$) to $d$-dimensional embeddings, i.e.,
\begin{equation}
\label{eq:dual-encoder}
\begin{aligned}
    \boldsymbol{x}=\text{Dual-Enc} \left(x; \theta \right) \in \mathbb{R}^{d},    
\end{aligned}
\end{equation}
where $\theta$ could be dense retriever ($\theta^\mathrm{den}$) or lexicon-aware retriever ($\theta^\mathrm{lex}$). With separately encoded query $\boldsymbol{q}$ and passage $\boldsymbol{p}$, we could calculate the relevance score via dot product for retrieval, i.e.,
\begin{equation}
\label{eq:score-function}
    \mathcal{R}(q, p; \theta)=\boldsymbol{q}^{T} \boldsymbol{p}.
\end{equation}
The dual-encoder architecture and lightweight dot product evaluation enable us to encode and index all passages in the collection $\mathcal{C}$ beforehand, so we only need to encode the given query for online retrieval, achieving more efficiency.

\paragraph{\textbf{Learning Framework for Retriever.}}
To train the dual-encoder $\theta$, we utilize contrastive learning following previous works~\cite{Xiong2021ANCE, Gao2021SimCSE}.
Specifically, with a given query $q$, a labeled positive passage $p^{+}$, and negative passages $\mathcal{N}$, contrastive loss can be applied to optimize the dual-encoder $\theta$ by maximizing the relevance of the $q$ and $p^{+}$ while minimizing that of $q$ and $p \in \mathcal{N}$, i.e.,
\begin{equation}
\label{eq:contrastive-learning}
    \mathcal{L}_{\theta}^{cl} = -\log \frac{\mathrm{exp}({\mathcal{R}(q, p^{+};\theta)})}
    {\sum_{p \in \{p^+\} \cap \mathcal{N}} \mathrm{exp}(\mathcal{R}(q, p;\theta))},
\end{equation}
where negative passage set $\mathcal{N}$ can be generated from top-ranked non-answer passages in retrieval results of BM25 model~\cite{Nguyen2016MSMARCO} or a trained retrievers~\cite{Zhan2021STAR-ADORE, Zhang2021AR2}, i.e.,
\begin{equation}
    \mathcal{N}=\left\{p \mid p \sim P\left(\mathcal{C} \backslash\left\{p^{+}\right\} \mid q ; \theta^{\mathrm{samp}}\right)\right\},
\end{equation}
where $P$ is a probability distribution over $\mathcal{C}$, which can be defined as non-parametric (e.g., $\theta^{\mathrm{samp}} = {\oslash}$) or parametric (e.g., $\theta^{\mathrm{samp}} \ne \oslash $). 

\paragraph{\textbf{Dense \& Lexicon-Aware Retrievers.}}
Both dense retriever ($\theta^{\mathrm{den}}$) and lexicon-aware retriever ($\theta^{\mathrm{lex}}$) follow dual-encoder architecture and the encoders are built upon PLMs like BERT~\cite{Devlin2019BERT}.
Precisely, a PLM ($\theta^{\mathrm{plm}}$) encodes a given text (i.e., query $q$ or passage $p$), $x = \{t_1, t_2, ...t_n\}$, to contextualized embeddings, i.e.,
\begin{equation}
\label{eq:PLM-encoding}
\begin{aligned}
    \boldsymbol{H}^x & = \text{PLM}(x; \theta^{\mathrm{plm}}) \\
    &= \text{PLM} \left([\mathrm{CLS}], t_1, t_2, ..., t_n, [\mathrm{SEP}]; \theta^{\mathrm{plm}}\right),
\end{aligned}
\end{equation}
eventually $\boldsymbol{H}^x = [\boldsymbol{h}_{[\mathrm{CLS}]}^x, \boldsymbol{h}_{1}^x, ..., \boldsymbol{h}_{n}^x, \boldsymbol{h}_{[\mathrm{SEP}]}^x]$. $[\mathrm{CLS}]$ and $[\mathrm{SEP}]$ are special tokens designed for sentence representation and separation by recent PLMs~\cite{Devlin2019BERT, Liu2019RoBERTa}. 
Dense retriever~\cite{Xiong2021ANCE, Qu2021RocketQA} represents text by using the embedding of special token $\mathrm{[CLS]}$ (i.e., $\boldsymbol{h}_{[\mathrm{CLS}]}^x$) as follows,
\begin{equation}
\label{eq:dense}
    \boldsymbol{x}^{\mathrm{den}} = \text{Dual-Enc}(x; \theta^{\mathrm{den}}) = \text{CLS-Pool}(\boldsymbol{H}^x),
\end{equation}
where $\theta^{\mathrm{den}} =\theta^{\mathrm{plm}}$ with no additional parameters.

For lexicon-aware retriever, we adopt SPLADE~\cite{Formal2021SPLADEv2} which learns to predict the weights of terms in PLM vocab for each token in the input $x$ by the Masked Language Modeling (MLM) layer and sparse regularization, then max-pooling these weights into a discrete text representation after log-saturation.
Formally, with $\boldsymbol{H}^x$ encoded by the PLM ($\theta^{\mathrm{plm}}$), a MLM layer ($\theta^{\mathrm{mlm}}$) linearly transform it into $\Tilde{\boldsymbol{H}}^x$, then term weight representation of $x$ could be obtained as follows,
\begin{equation}
\label{eq:splade}
\begin{aligned}
    \boldsymbol{x}^{\mathrm{lex}} & = \text{Dual-Enc}(x; \theta^{\mathrm{lex}}) \\
                & = \text{MAX-Pool}(\log(1 + \text{ReLU}(\boldsymbol{W}^e \cdot \Tilde{\boldsymbol{H}}^x))),
\end{aligned}
\end{equation}
where $\boldsymbol{W}^e \in \mathbb{R}^{|V|\times e}$ is the transpose of the input embedding matrix in PLM as the MLM head, and the $\theta^{\mathrm{lex}} = \{ \theta^{\mathrm{plm}}, \theta^{\mathrm{mlm}}, \boldsymbol{W}^e\}$.

The dense encoder represents texts as global sequence-level embeddings and is good at global semantic matching, while the lexicon-aware encoder represents local term-level embeddings and handles salient phrases and entity mentions well.
Both encoders can be optimized with the Eq.~\ref{eq:contrastive-learning}.

\subsection{Lexical Enlightened Dense Retriever} \label{sec:LED}
Fig.~\ref{fig:training-process} illustrates the training workflow of our LED retriever.
Specifically, we follow a two-stage training procedure. In the \textit{Warmup} stage, we independently train the dense and lexicon-aware retrievers by Eq.~\ref{eq:contrastive-learning}, both with BM25 negatives ($\mathcal{N}^{\mathrm{bm25}}$).
This stage ends up with two retrievers, namely the Lexical Warm-up ($\theta^{\mathrm{lex1}}$) and the Dense Warm-up ($\theta^{\mathrm{den1}}$).

Then, we sample negative passages ($\mathcal{N}^\mathrm{lex1}$) with the Lexcial Warm-up checkpoint ($\theta^{\mathrm{lex1}}$) for the second stage, namely \textit{Continual Training} stage. With the fixed negative passages ($\mathcal{N}^\mathrm{lex1}$), we continually train the lexical retriever initialized from the warm-up checkpoint ($\theta^{\mathrm{lex1}}$) by Eq.~\ref{eq:contrastive-learning}. After the second stage, we could obtain the model named Lexical ($\theta^{\mathrm{lex2}}$), which plays a role of a teacher for later lexical knowledge teaching.

With a well-trained lexicon-aware teacher ($\theta^{\mathrm{lex2}}$) and dense student after warming up ($\theta^{\mathrm{den1}}$), we enlighten the student by transferring knowledge from the teacher. The knowledge transfer is achieved from two perspectives -- 1) a lexicon-augmented contrastive objective to challenge the dense encoder and 2) a rank-consistent regularization to make the dense model's behavior inclined to its lexicon-aware teacher. 
We will detail the two objectives in the following paragraphs.

\paragraph{\textbf{Lexicon-Augmented Contrastive.}}
Following previous work~\cite{Zhan2021STAR-ADORE}, we use the dense negatives ($\mathcal{N}^\mathrm{den1}$) sampled by the Dense Warm-up ($\theta^{\mathrm{den1}}$) to boost dense retrieval.
Meanwhile, inspired by~\citet{Chen2021ImitateSparse} who trained a lexical retriever with negatives provided by term-based techniques such as BM25, we try to enhance the dense retriever from the negatives augmentation perspective.

Considering the similar backbone (i.e., both are fine-tuned on PLMs) and the same optimization objectives (i.e., Eq.~\ref{eq:contrastive-learning}) of both dense and lexicon-aware retrievers, their significant differences in retrieval behaviors may partially stem from training with different negative passages. So intuitively, dense one can use the lexical negatives ($\mathcal{N}^\mathrm{lex1}$) to partially imitate the training process of the lexical teacher ($\theta^{\mathrm{lex2}}$), thus learning lexicon-aware ability. 
One step further, we use the lexical negatives ($\mathcal{N}^\mathrm{lex2}$) for learning more and harder lexical knowledge. 
Meanwhile, compared to the dense negatives ($\mathcal{N}^\mathrm{den1}$), these lexical negatives ($\mathcal{N}^\mathrm{lex1}$ and $\mathcal{N}^\mathrm{lex2}$) can provide more diverse examples and could be regarded as adversarial examples to challenge the dense retriever for robust retriever training.
Formally, the lexicon-augmented contrastive loss for LED is,
\begin{equation}
\label{eq:contrastive-learning-lex}
    \mathcal{L}_{\theta^{\mathrm{led}}}^{cl} = -\log \frac{\mathrm{exp}({\mathcal{R}(q, p^{+};\theta^{\mathrm{led}})})}
    {\sum_{p \in \{p^+\} \cap \mathcal{N}^{\mathrm{mix}}} \mathrm{exp}(\mathcal{R}(q, p;\theta^{\mathrm{led}}))},
\end{equation}
where $\mathcal{N}^{\mathrm{mix}}=\{\mathcal{N}^\mathrm{lex1} \cap \mathcal{N}^\mathrm{lex2} \cap \mathcal{N}^\mathrm{den1}\}$.

\paragraph{\textbf{Rank-Consistent Regularization.}}

From the retrieval behavior perspective, for given query-passage pairs, {we utilize the lexicon-aware teacher ($\theta^{\mathrm{lex2}}$) to generate ranking pairs to regularize and guide LED's retrieval behavior.}

Specifically, given a query $q$ and passages from $\mathcal{D}^q = \{p \in \{p^+\}  \cap \mathcal{N}^{\mathrm{mix}} \}$,
the Lexical ($\theta^{\mathrm{lex2}}$) scores each query-passage pair (abbr. $\mathcal{R}(p;\theta^\mathrm{lex2})$) with Eq.~\ref{eq:score-function} and generate ranking pairs as follows,
\begin{equation}
    \small
    \mathcal{K}^q = \left\{(p_i, p_j) | 
            p_i, p_j \in \mathcal{D}^q,
            \mathcal{R}(p_i; \theta^{\mathrm{lex2}}) > \mathcal{R}(p_j; \theta^{\mathrm{lex2}})
        \right\}.
\end{equation}
Then, a pair-wise rank-consistent regularization is employed to make the dense model’s behavior incline to the lexicon-aware one by minimizing the following margin-based ranking loss,
\begin{equation}
\label{eq:lexical-learning}
    \small
    \mathcal{L}^{ll}_{\theta^{\mathrm{led}}} = \frac{1}{\mid \mathcal{K}^q \mid} \sum_{ p_i, p_j \in \mathcal{K}^q} \max[0, \mathcal{R}(p_i; \theta^{\mathrm{led}}) - \mathcal{R}(p_j; \theta^{\mathrm{led}})],
\end{equation}
where $\mathcal{R}(p; \theta^{\mathrm{led}})$ is the abbrivation of relevance $\mathcal{R}(q, p; \theta^{\mathrm{led}})$ calculated by LED ($\theta^{\mathrm{led}}$) with Eq.~\ref{eq:score-function}.
Compared to logits distillation with a list-wise KL-divergence loss, our training objective provides a weak supervision signal pair-wisely, thus keeping the effects of injecting lexical knowledge on dense properties at a minimum level. Especially since we don't punish the dense student as long as its ranking of a given pair is the same as the teacher, without strict requirements on the score gap like Margin-MSE~\cite{Hofstatter2020Margin-MSE}. Experiments in Tab.~\ref{tab:distillation} demonstrate the merit of our method.

\paragraph{\textbf{Training and Inference.}}
To incorporate lexicon-aware ability while keeping its sequence-level semantic representation ability for passage retrieval, we combine contrastive loss ($\mathcal{L}^{cl}_{\theta^{\mathrm{led}}}$) in Eq.~\ref{eq:contrastive-learning-lex} and lexical learning loss ($\mathcal{L}^{ll}_{\theta^{\mathrm{led}}}$) in Eq.~\ref{eq:lexical-learning} to train our LED retriever ($\theta^{\mathrm{led}}$) as follows,
\begin{equation}
    \label{eq:final-led-loss}
    \mathcal{L}_{\theta^{\mathrm{led}}} = \mathcal{L}^{cl}_{\theta^{\mathrm{led}}} + \lambda \mathcal{L}^{ll}_{\theta^{\mathrm{led}}},
\end{equation}
where $\lambda$ is a hyperparameter to control how intensive the training inclines to transfer lexical-ware knowledge from the lexicon-aware teacher ($\theta^{\mathrm{lex2}}$).

For inference, LED pre-computes the embeddings of all passages in the entire collection $\mathcal{C}$ and builds indexes with FAISS~\cite{John2021ANN}. Then LED encodes queries online and retrieves top-ranked $k$ passages based on the relevance score.

{\paragraph{\textbf{Remark}}
Our framework injects lexicon-aware capability into a sequence-level representation model, showing two-fold advantages in comparison to previous methods superficially combining dense and lexicon-aware retrievers: 
1) Compared to using two separate PLM-based dense and lexicon-aware retrievers~\cite{Chen2021ImitateSparse, Shen2022UnifieR, Lin2021inbatch, Lin2021UniCOIL}, our LED retriever could achieve hybrid retrieval results and comparable performances with only one model. We no longer need to maintain multiple index systems for both retrieval models in an ensemble system or to encode an online query twice with different retrievers, reducing memory footprint and inference time. 
2) Compared to fusing PLM-based dense and traditional term-matching retrievers like BM25~\cite{Li2021Pseudo, Li2022Interpolate,Chen2021ImitateSparse, Lin2021inbatch, Lin2021UniCOIL, Seo2019DENSPI, Ma2021QGenHyb, Luan2021MEBERT, Wang2021Interpolation}, our single method could achieve better results since the lexicon-aware capability of LED is learned from a strong retriever (shown in Tab. ~\ref{tab:ensemble}).}

\section{Experiments}

\begin{table*}[ht]
\small
\centering

\caption{Experimental results on MS MARCO, TREC DL 2019 (DL'19), and TREC DL 2020 (DL'20) datasets (\%). We mark the best results in \textbf{bold} and the second-best underlined.  Numbers marked with `$\mbox{*}$' mean that the improvement is statistically significant compared with the baseline (t-test with $p$-value $<0.05$).
}

\resizebox{0.88\textwidth}{!}{
\begin{tabular}{llccccccc}
\toprule
\multicolumn{1}{c}{\multirow{2}{*}{\textbf{Methods}}} & \multicolumn{1}{c}{\multirow{2}{*}{\textbf{PLM}}} &
\multicolumn{1}{c}{\multirow{2}{*}{\textbf{Ranker}}} & \multicolumn{1}{c}{\multirow{2}{*}{\textbf{Multi}}} &
\multicolumn{3}{c}{\textbf{MS MARCO Dev}}     & \multicolumn{1}{c}{\textbf{DL'19}} & \multicolumn{1}{c}{\textbf{DL'20}} \\ \cmidrule(lr){5-7} \cmidrule(lr){8-8} \cmidrule(l){9-9} 
\multicolumn{1}{c}{}   & \multicolumn{1}{c}{}                                &   \textbf{Taught}      & \textbf{Vector} & MRR@10        & R@50          & R@1k          & NDCG@10            & NDCG@10            \\ \hline \hline

\multicolumn{9}{l}{\textit{\textbf{Lexicon-Aware Retriever}}}               \\ \hline

BM25~\cite{Robertson2009BM25}                         & -                     &                &                & 18.7          & 59.2          & 85.7          & 50.6           & 48.0        \\
DeepCT~\cite{Dai2019DeepCT}                           & BERT$_\text{base}$    &                &                & 24.3          & 69.0          & 91.0          & 55.1           & 55.6        \\
COIL-full~\cite{Gao2021COIL}                          & BERT$_\text{base}$    &                &                & 35.5          & -             & 96.3          & 70.4           & -           \\
UniCOIL~\cite{Lin2021UniCOIL}                         & BERT$_\text{base}$    &                &                & 35.2          & 80.7          & 95.8          & -              & -           \\
SPLADE-max~\cite{Formal2021SPLADEv2}                  & DistilBERT            &                &                & 34.0          & -             & 96.5          & 68.4           & -           \\
DistilSPLADE-max~\cite{Formal2021SPLADEv2}            & DistilBERT            &  $\checkmark$  &                & 36.8          & -             & 97.9          & \uline{72.9}   & -           \\
UniCOIL $\Lambda$~\cite{Chen2021ImitateSparse}        & BERT$_\text{base}$    &                &                & 34.1          & 82.1          & 97.0          & -              & -           \\ \hline \hline

\multicolumn{9}{l}{\textit{\textbf{Dense Retriever}}}                        \\ \hline
ANCE~\cite{Xiong2021ANCE}                             & RoBERTa$_\text{base}$ &                &                & 33.0          & -             & 95.9          & 64.5               & 64.6              \\
ADORE~\cite{Zhan2021STAR-ADORE}                       & RoBERTa$_\text{base}$ &                &                & 34.7          & -             & -             & 68.3               & 66.6              \\
TAS-B~\cite{Hofstatter2021TAS-B}                      & DistilBERT            &  $\checkmark$  &                & 34.7          & -             & 97.8          & 71.7               & 68.5      \\
TAS-B + CL-DRD~\cite{Zeng2022CL-DRD}             & DistilBERT            &  $\checkmark$  &                & 38.2          & -             & -             & 72.5               & 68.7      \\
TCT-ColBERT~\cite{Lin2021inbatch}                     & BERT$_\text{base}$    &  $\checkmark$  &                & 35.9          & -             & 97.0          & 71.9               & -                 \\
ColBERTv1~\cite{Khattab2020ColBERT}                   & BERT$_\text{base}$    &                &   $\checkmark$ & 36.0          & 82.9          & 96.8          & 67.0               & 66.8              \\
ColBERTv2~\cite{Khattab2021ColBERTv2}                 & BERT$_\text{base}$    &  $\checkmark$  &   $\checkmark$ & \uline{39.7}  & \uline{86.8}  & \textbf{98.4} & 72.0          & 62.1              \\
coCondenser~\cite{Gao2021coCondenser}                 & BERT$_\text{base}$    &                &                & 38.2          & -             & \textbf{98.4} & -               & -                 \\
PAIR~\cite{Ren2021PAIR}                               & ERNIE$_\text{base}$   &  $\checkmark$  &                & 37.9          & 86.4          & 98.2          & -                  & -                 \\
RocketQAv2~\cite{Ren2021RocketQAv2}                   & ERNIE$_\text{base}$   &  $\checkmark$  &                & 38.8          & 86.2          & 98.1          & -                  & -                 \\
AR2-G~\cite{Zhang2021AR2}                             & BERT$_\text{base}$    &  $\checkmark$  &                & 39.5          & -             & -             & -                  & -                 \\
\hline \hline
\multicolumn{8}{l}{\textit{\textbf{Our Models}}}             \\ \hline
LEX (Warm-up)                                      & DistilBERT             &                &                & 36.1          & 84.2          & 97.5          & 67.4               & 66.4               \\ 
LEX (Continue)                                     & DistilBERT             &                &                & 38.3          & 85.9          & 98.0          & 72.8               & 67.7               \\ \hline
DEN (Warm-up)                                      & BERT$_\text{base}$     &                &                & 36.1          & 83.5          & 97.7          & 64.7               & 65.9               \\
DEN (Continue)                                     & BERT$_\text{base}$     &                &                & 38.1          & 86.3          & \textbf{98.4} & 69.1               & 67.8               \\
DEN (w/ RT)                                        & BERT$_\text{base}$     &  $\checkmark$  &                & 39.6          & 86.7          & \textbf{98.4} & 71.8               & \uline{69.7}       \\
\midrule 
\rowcolor{LightCyan}
LED                                                  & BERT$_\text{base}$     &                &                & 39.6          		& 86.6          & \uline{98.3}  & 70.5             & 67.9               \\
\rowcolor{LightCyan}
LED (w/ RT)                                          & BERT$_\text{base}$     &  $\checkmark$  &                & \textbf{40.2}$^{*}$ & \textbf{87.6}$^{*}$  & \textbf{98.4} & \textbf{74.4}$^{*}$ & \textbf{70.2}$^{*}$ \\ 
\bottomrule

\end{tabular}
}
\label{tab:main-table}
\end{table*}

We evaluate our retriever on three public human-annotated real-world benchmarks, namely MS MARCO~\cite{Nguyen2016MSMARCO}, TREC Deep Learning 2019~\cite{Craswell2020TREC19}, and TREC Deep Learning 2020~\cite{Craswell2020TREC20}.
MS-MARCO Dev has 6980 queries, TREC 2019 has 43 queries, and TREC 2020 includes 54 queries. 
In all three benchmarks, first-stage retrievers are required to fetch relevant passages from an 8-million scale collection.
We report MRR@10, Recall@50, and Recall@1000 for MS MARCO Dev, as well as NDCG@10 for both TREC Deep Learning 2019 and TREC Deep Learning 2020. For all three datasets, we use the official TREC evaluation files to conduct the evaluation protocol.\footnote{https://github.com/usnistgov/trec\_eval}

\subsection{Baselines}
We compare with previous state-of-the-art baselines including traditional term-based techniques like BM25~\cite{Robertson2009BM25}, and dense~\cite{Xiong2021ANCE,Zhan2021STAR-ADORE,Hofstatter2021TAS-B,Zeng2022CL-DRD,Khattab2020ColBERT,Khattab2021ColBERTv2,Lin2021inbatch,Gao2021coCondenser,Ren2021PAIR,Ren2021RocketQAv2,Zhang2021AR2} as well as lexicon-aware retrievers~\cite{Dai2019DeepCT,Gao2021COIL,Lin2021UniCOIL,Formal2021SPLADEv2,Chen2021ImitateSparse}. 
More details about the baselines are provided in Appendix \ref{appendix-sec:baselines}.
{We report the models used during the two-stage training pipeline for a detailed comparison. 
For lexicon-aware retrievers, we report the models after the warm-up training, namely LEX (Warm-up), and continual training, namely LEX (Continue). Note that LEX (Continue) is the lexical teacher used for teaching. Similarly, for dense retrievers, we show DEN (Warm-up) and DEN (Continue). Note that the DEN (Warm-up) is the dense student where our LED model starts from and DEN (Continue) is independently trained with the hard negatives provided by DEN (Warm-up) by Eq.~\ref{eq:contrastive-learning}. We also report the DEN (Continue) further enhanced with a strong ranker distillation (i.e., DEN (w/ RT)).}

\subsection{Implementation Details}
\label{appendix-sec:implementation}
All experiments run on $1$ NVIDIA Tesla A100 GPU having 80GB memories with a fixed random seed. We train our models with mixed precision to speed up the training and meet the huge memory need. The training time will last about 32 hours.
For the lexical teacher, we train a DistilBERT~\cite{Sanh2019DistilBERT} following SPLADE-max~\cite{Formal2021SPLADEv2}.\footnote{https://huggingface.co/distilbert-base-uncased} 
Following previous work~\cite{Zhou2022R2anker}, in the warm-up stage, we train the lexical retriever with batch size $48$, $5$ negatives for each query randomly sampled from BM25 negatives, and a learning rate $3e^{-5}$ for three epochs. 
In the second stage, we remain all hyperparameters unchanged except lower the learning rate to $2e^{-5}$ and use negative passages randomly sampled from the top $200$ self-mining hard negatives.
For dense retriever, we train coCondenser~\cite{Gao2021coCondenser} checkpoint with batch size $16$, $7$ negatives per query, and a learning rate of $1e^{-5}$ for three epochs.\footnote{https://huggingface.co/Luyu/co-condenser-marco. We chose the coCondenser as the dense retriever due to its retrieval-oriented pre-training and its superior performance compared to vanilla BERT and DistilBERT checkpoints. We tested our proposed strategies on these models as well and saw similar improvements as with coCondenser.}

Particularly, in the LED training stage, with other hyperparameters unchanged, we set the learning rate $5e^{-6}$ and randomly select $32$ hard negatives from the mixture of each top $200$ negatives mined by the warm-up dense student, warm-up lexical teacher, and final lexical teacher. The number of negatives per query $32$ is selected from $\{8, 16, 24, 32\}$. The higher number of negatives per query indicates the more pair-wise ranks constructed by the lexical teacher, leading to more lexical knowledge transfer.

For rank-consistent regularization, 
we set loss weight $\lambda = 1.2$ after searching from $\{1.0, 1.2, 1.5, 1.8, 2.0\}$.


\subsection{Main Results}
As shown in Tab.~\ref{tab:main-table}, we present the evaluation results on the aforementioned three public benchmarks.

Firstly, our LED retriever achieves comparable performance with state-of-the-art methods ColBERTv2~\cite{Khattab2021ColBERTv2} and AR2~\cite{Zhang2021AR2} on MS MARCO Dev, although both baselines are taught by the powerful ranking model (i.e., cross-encoder). 
After coupling with a similar ranker distillation, our LED retriever (i.e., LED (w/ RT)) can be further improved and meanwhile outperforms state-of-the-art baselines on all three datasets, showing the compatibility of distillation from lexicon-aware sparse retriever.\footnote{Like AR2~\cite{Zhang2021AR2} and ColBERTv2~\cite{Khattab2021ColBERTv2}, we use KL-divergence to distill the ranker’s scores into the LED model, but we use ERNIE-2.0-base~\cite{Sun2020ERNIE2.0} instead of ERNIE-2.0-large in AR2. The KL loss is directly added to Eq.~\ref{eq:final-led-loss} during the training of LED.}
Note that we neither use heavy ranker teacher in AR2~\cite{Zhang2021AR2} nor multiple vector representation applied in ColBERTv2~\cite{Khattab2021ColBERTv2}.

Secondly, LED (w/ RT) achieves better performance than DEN (w/ RT) on all three datasets, demonstrating that our training method can transfer some complementary lexicon-aware knowledge not covered by the cross-encoder. 
The weak intensity of the supervision signal makes our lexical-enlightened strategy a promising plug-and-play technique for other dense retrievers.

Thirdly, our LED retriever taught by a smaller lexicon-aware retriever is similarly performant as the dense retriever taught by a strong cross-encoder (i.e., DEN (w/ RT)), showing the effectiveness of injecting lexical knowledge into the dense retriever.
The reasons are two-fold: (1) The dual-encoder architecture of the lexicon-aware teacher enables the relevance calculation can be easily integrated into in-batch techniques to scale up the teaching amount. (2) More importantly, lexicon-aware retriever could provide self-mining hard negatives for more direct supervision while cross-encoder can only provide score distribution over given passages.


\subsection{Further Analysis}
\paragraph{\textbf{Comparison of Teaching Strategies.}}

\begin{table}[t!]
\centering
\small
\caption{Evaluation results of different teaching strategies on MS MARCO Dev (\%). 
`$\mbox{*}$' refers to statistical significance.
}
\resizebox{0.32\textwidth}{!}{
\begin{tabular}{lcc}
\toprule
\multicolumn{1}{c}{\textbf{Methods}}        & \textbf{MRR@10} & \textbf{R@1k} \\ \midrule
No Distillation                             & 38.1            & \textbf{98.4} \\ \midrule
Filter~\cite{Qu2021RocketQA}                & 38.4            & \textbf{98.4} \\ 
Margin-MSE~\cite{Hofstatter2020Margin-MSE}  & 38.5            & 98.3          \\
ListNet~\cite{Xiao2022Distill-VQ}           & 38.7            & 98.2          \\
KL-Divergence~\cite{Zhang2021AR2}           & 39.0            & \textbf{98.4} \\ \midrule
Ours                                        & \textbf{39.6$^*$}   & 98.3 \\ \bottomrule
\end{tabular}
}
\label{tab:distillation}
\end{table}

Tab.~\ref{tab:distillation} shows the comparison of our proposed pair-wise rank-consistent regularization with other teaching strategies. Filter means the negatives with high scores (i.e., false negatives) are filtered by LEX (Continue). The other three strategies (e.g., Margin-MSE, ListNet, and KL-Divergence) are borrowed from knowledge distillation in IR domain.
From the table, we can find that all strategies can bring performance gain, even in an indirect way like Filter. This observation proves that learning from the lexicon-aware representation model leads to a better dense retriever. Also, our rank-consistent regularization outperforms other baselines on MRR@10 metric by a large margin, showing the superiority of our method.
Besides, we can find that the point-wise objective (i.e., Margin-MSE) brings the least gain, followed by the list-wise objectives (i.e., ListNet and KL-Divergence) and our pair-wise rank-consistent regularization brings the most significant gain.
The phenomenon implies that a soft teaching objective is more functional for transferring knowledge from the lexicon-aware model than strict objectives.
In fact, enforcing dense retrievers to be aligned with fine-grained differences between scores of the LEX often leads to training collapse. 
Concretely, only equipped with carefully chosen hyperparameters, especially small distillation loss weight, Margin-MSE can enhance the dense retriever.

\paragraph{\textbf{Comparison of Ensemble Retrievers.}}
\begin{table}[t!]
\small
\centering
\caption{Comparison with Ensemble Systems on MS MARCO Dev (\%). The first block results are copied from~\cite{Lin2021inbatch, Lin2021UniCOIL,Chen2021ImitateSparse}. $\Lambda$~\cite{Chen2021ImitateSparse} refers to a dense retriever trained with data generated by lexicon-based methods such as BM25 and UniCOIL. `$\mbox{*}$' indicates statistical significance compared to their counterparts without our training strategies.}
\resizebox{\linewidth}{!}{
\begin{tabular}{lccc}
\toprule
\multicolumn{1}{c}{\textbf{Ensemble Systems}} & \textbf{MRR@10} & \textbf{R@50} & \textbf{R@1k} \\ \midrule
TCT-ColBERT + BM25~\cite{Lin2021inbatch}                  & 36.9            & -             & -             \\
TCT-ColBERT + UniCOL~\cite{Lin2021UniCOIL}                & 37.8            & -             & -             \\
TCT-ColBERT + UniCOL~\cite{Lin2021UniCOIL}                & 38.2            & -             & -             \\
ANCE + BM25~\cite{Chen2021ImitateSparse}                  & 34.7            & 81.6          & 96.9          \\
RocketQA + BM25~\cite{Chen2021ImitateSparse}              & 38.1            & 85.9          & 98.0          \\
RocketQA + UniCOIL~\cite{Chen2021ImitateSparse}           & 38.8            & 86.5          & 97.3          \\
RocketQA + BM25 $\Lambda$~\cite{Chen2021ImitateSparse}    & 37.9            & 85.7          & 98.0          \\
RocketQA + UniCOIL $\Lambda$~\cite{Chen2021ImitateSparse} & 38.6            & 86.3          & 98.5          \\ \midrule
DEN (Continue) + BM25 & 30.4            & 87.1          & 98.6          \\
DEN (Continue) + LED	                         & 39.3	           & 86.9	       & 98.5          \\
DEN (Continue) + DEN (w/ RT)	                     & 39.4	           & 87.0	       & 98.5          \\
DEN (Continue) + LEX (Continue)                           & 40.4            & 88.4          & \textbf{98.7} \\
DEN (w/ RT) + LEX (Continue)                     & 40.7	           & 88.4	       & \textbf{98.7} \\
LED + LEX (Continue)                           & \textbf{40.9$^*$}   & 88.3          & 98.6          \\
LED (w/ RT) + LEX (Continue)                     & \textbf{41.1$^*$}   & \textbf{88.5} & \textbf{98.7} \\ \bottomrule
\end{tabular}
}

\label{tab:ensemble}
\end{table}

We are also curious whether our LED can improve the performance of ensemble retrievers. With LEX (Continue) ($\theta^{\mathrm{lex2}}$) and LED ($\theta^{\mathrm{led}}$), we simply use the summation of the normalized relevance scores of two retrievers, and then return a new order of retrieval results.
Tab.~\ref{tab:ensemble} gives the evaluation results of our systems and other strong baselines reported in previous work~\cite{Chen2021ImitateSparse, Lin2021inbatch, Lin2021UniCOIL}. {Note that previous work~\cite{Lin2021inbatch, Lin2021UniCOIL} utilized weighted score sum after hyper-parameter searching while we directly sum the normalized scores of two retrievers without any tuning.} From the results in Tab.~\ref{tab:ensemble}, we can observe:

(1) Aligned with results in SPAR~\cite{Chen2021ImitateSparse}, the ensemble of two dense retrievers (i.e., DEN (Continue) + LED and DEN (Continue)  + DEN (w/ RT)) is not as performant as that of one dense and one lexicon-aware retriever.
In particular, the ensemble of two dense retrievers is even less competitive than a single LED or DEN (w/ RT). The results are rational because two base models have similar retrieval behaviors and the strong one will be impeded by the weak one if they have the same weight in the ensemble system. 
{The latter reason could also be used to explain why the ensemble of dense and the traditional term-based technique like BM25 (i.e., DEN (Continue) + BM25) is less good than the single DEN (Continue).}

(2) Although coupling with LEX (Continue) will not introduce new knowledge to the hybrid ensemble system where LED is the base retriever, LED + LEX (Continue) can further boost the performance of DEN (Continue) + LEX (Continue).
The reason behind this is that the LED scores golden query-passage pairs higher than DEN, so these pairs are ranked higher in the later ensemble process.
This behavior could be regarded as an instance-level weighted score aggregation inside the network and it is more feasible to obtain than tuning the weights of retrievers for each query in the ensemble system. {This observation could from the side prove that our dense and lexicon-aware abilities fusion inside the network is better than a superficial ensemble.}

(3) LED (w/ RT) + LEX (Continue) is slightly better than LED + LEX (Continue) and DEN (w/ RT) + LEX (Continue), once again proving that our lexical rank-consistent regularization is complementary to the ranker distillation.

\paragraph{\textbf{Impact of Different Components.}}

\begin{table}[t!]
\small
\centering
\caption{Ablation Study on MS MARCO Dev (\%). Negs is short for negatives. 
`$\mbox{*}$' indicates statistical significance.
}
\resizebox{0.88\linewidth}{!}{
\begin{tabular}{lcc}
\toprule
\multicolumn{1}{c}{\textbf{Retrievers}} & \textbf{MRR@10} & \textbf{R@1k} \\ \midrule
LED                                  & \textbf{39.6$^*$}   & 98.3          \\ \midrule
\ \ \ w/o Rank Regularization      & 37.9          & \textbf{98.5} \\
\ \ \ w/o LEX Continue Negs ($\mathcal{N}^{\mathrm{lex2}}$)   & 39.4          & 98.3          \\
\ \ \ w/o LEX Warm-up Negs ($\mathcal{N}^{\mathrm{lex1}}$)      & 39.4          & 98.3          \\ 
\ \ \ w/o LEX Mixed Negs ($\mathcal{N}^{\mathrm{lex1}}$ $\cap$ $\mathcal{N}^{\mathrm{lex2}}$)              & 39.2          & 98.4          \\\bottomrule
\end{tabular} 
}

\label{tab:ablation}
\end{table}
We conduct an ablation study to further investigate the impact of lexical hard negatives and rank-consistent regularization method. Tab.~\ref{tab:ablation} reports the results of removing each component.
We can observe that pair-wise rank-consistent regularization plays an important role in lexical learning because removing it will bring significant performance degradation on MRR@10 metric.
In addition, we can find that both negatives provided by LEX (Warm-up) and LEX (Continue) are both helpful for the contrastive training of the dense retriever, and removing both of them results in a more obvious performance drop.

\begin{figure}[t!]
\centering
\subfigure[Top-ranked samples]{
\label{fig:top-dist-0.2}
\includegraphics[scale=0.30]{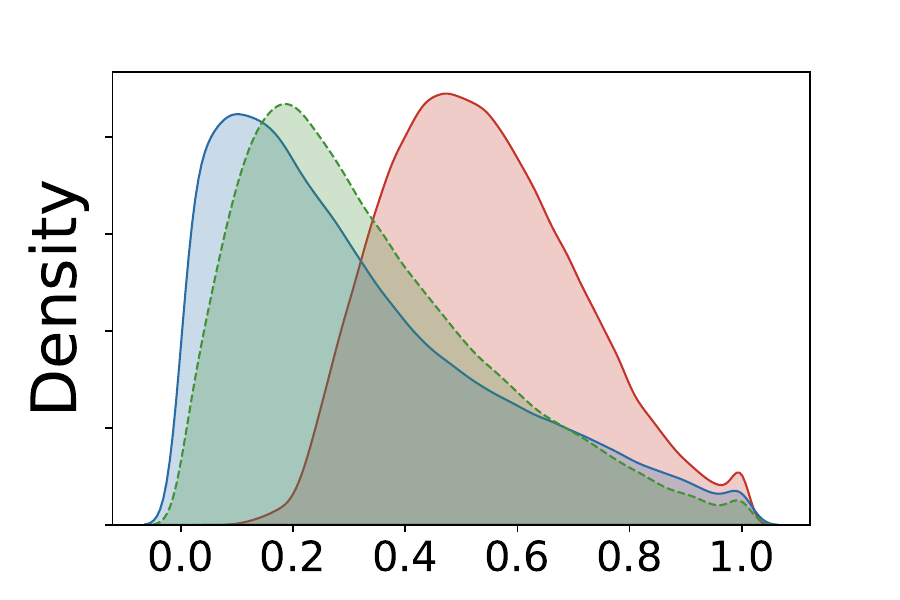}}
\hspace{-10mm}
\subfigure[Bottom-ranked samples]{
\label{fig:bot-dist-0.2}
\includegraphics[scale=0.30]{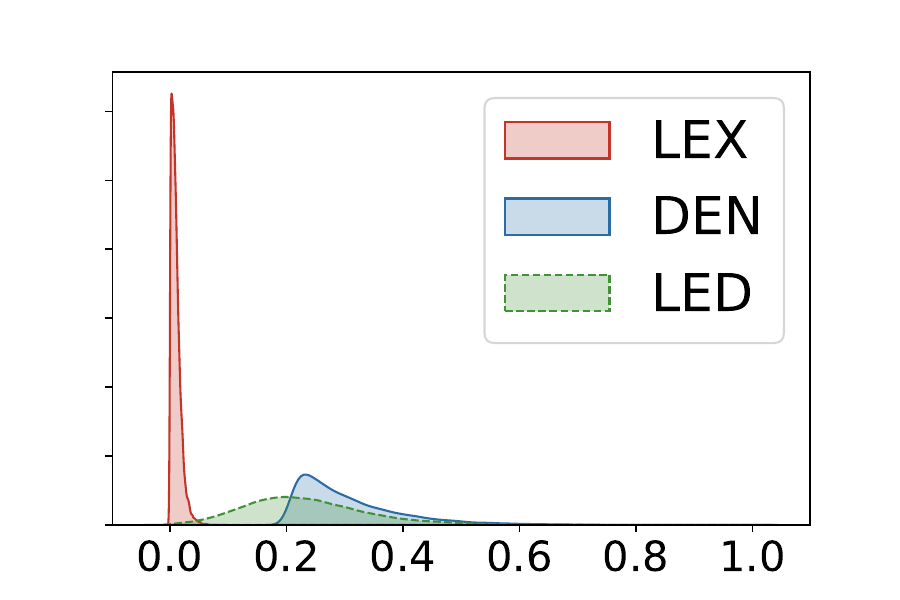}}
\caption{ Distributions of model prediction for DEN (Continue), LEX (Continue), and LED retrievers over MS MARCO Dev. 
For visual clarity, we use the query-passage pairs which the LEX and DEN predict discrepantly as data samples. 
The discrepancy is determined by that there is a $>0.2$ margin between their predicted scores normalized over passages retrieved for a $q$. To ensure diversity, we consider two normalization cases, LEX-biased pairs (i.e., LEX's top-100) and LEX-unbiased pairs (i.e., LEX's bottom-100 out of 1000).
}
\label{fig:distribution-shift}
\end{figure}

\paragraph{\textbf{Effects on Model Predictions.}}
\label{sec:distribution-shift}
To further check the effects of learning lexicon-aware capability on the LED, we illustrate the distribution shift of predictions of dense retrievers before and after lexical enlightenment in Fig.~\ref{fig:distribution-shift}. We can make the following observations:
(1) In both two sets of query-passage pairs, compared to DEN distributions, the score distributions of LED are clearly shifting to the LEX, showing the success of lexical knowledge learning.
(2) LED's distribution remains more overlaps with DEN instead of LEX, which proves that our rank-consistent regularization method could keep LED's dense retriever properties, thanks to the weak supervision signal.

\begin{figure}[t!]
\centering
\subfigure[Effects of \#Negatives]{
\label{fig:NN-impact}
\includegraphics[scale=0.235]{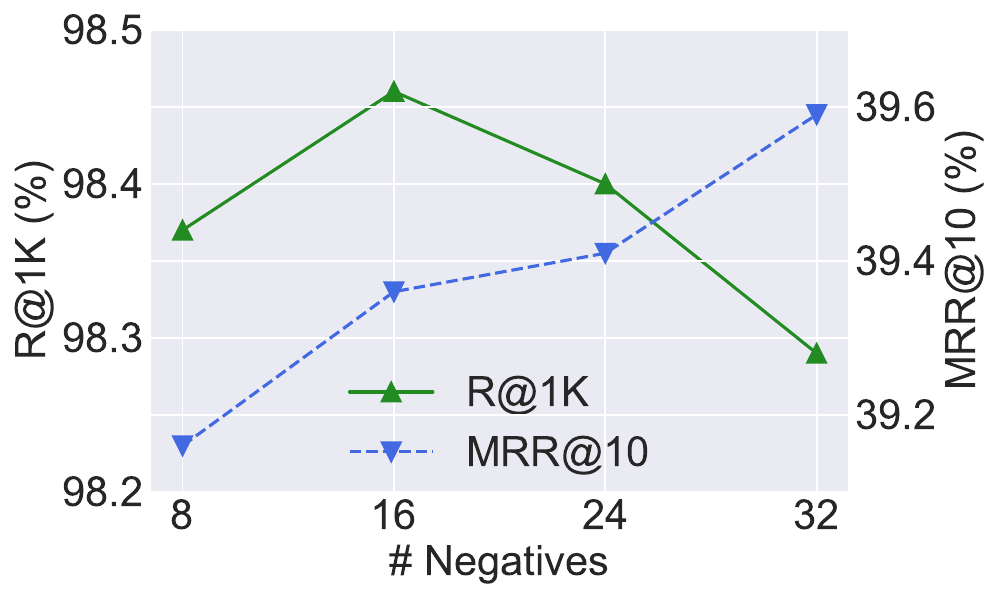}}
\hspace{-2mm}
\subfigure[Effects of $\lambda$]{
\label{fig:weight-impact}
\includegraphics[scale=0.235]{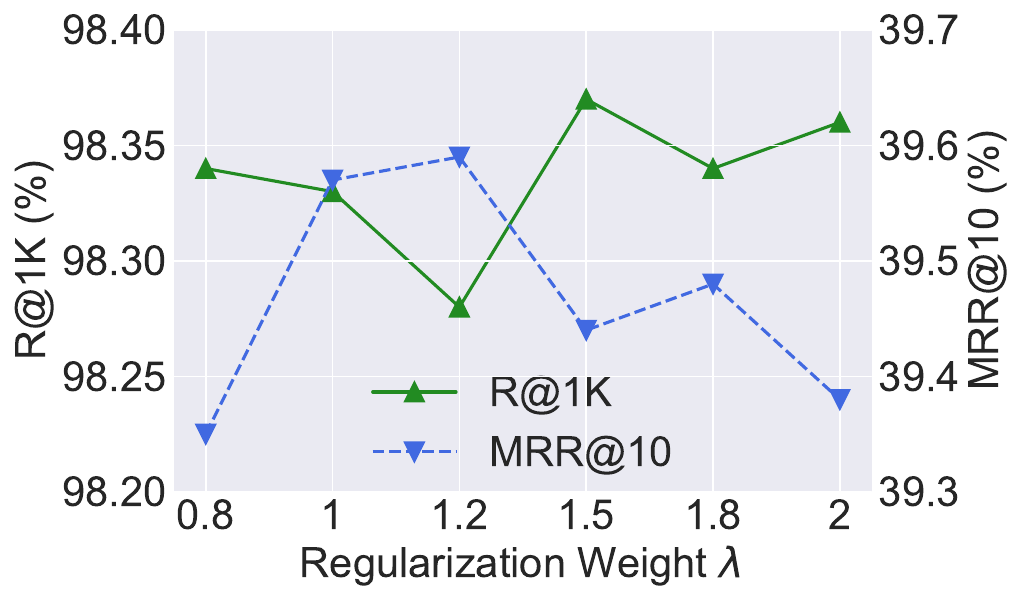}}
\caption{(a) Effects of the number of negatives per query on MS MARCO Dev. (b) Effects of the regularization weight $\lambda$ on MS MARCO Dev.}
\label{fig:distribution-shift}
\end{figure}

\paragraph{\textbf{Impacts of Hyperparameters}}
We conduct extra experiments to explore the impact of hyperparameters on LED retriever training.
Fig.~\ref{fig:NN-impact} illustrates the impact of changing negative passages on the LED.
We can observe that as the number of negative passages increases, the MRR@10 performance goes up and the R@1k performance reaches the peak when $16$ and decreases gradually.
The main table shows that Lexical is less performant than Dense at R@1k metric (98.0 $<$ 98.4).
So the trend of increasing the number of negatives proves that imitating too much the lexical retriever will also be negatively influenced by the weakness of the teacher.
These two trends indicate that, with more negatives, the teacher will construct more rank pairs for more lexical knowledge transfer.
Fig.~\ref{fig:weight-impact} shows the performance with regard to different regularization weights $\lambda$. It is observed that the performances don't fluctuate significantly as the weight $\lambda$ changes, demonstrating the robustness of lexical enhancement strategies.
Interestingly, the increase in MRR@10 comes with the drop in R@1k to some extent, once again showing that a well-enhanced LED also inherits the weakness of Lexical.

\paragraph{\textbf{Zoom-in Study of Retrieval Ranking.}}
\label{sec:zoom-in-ranking-change}
Tab.~\ref{tab:avg-rank} shows how the average rank of golden passages varies across different rank ranges, bucketed by LEX-predicted ranks of the golden passages. We can observe that: (1) More than 50\% golden passages are ranked in the top 5 by the LEX, paving the way for good lexical teaching.
(2) The average ranking of golden passages by LED is consistently improved until the top 100, which means approximately 90\% of answers are ranked higher by the retriever after learning lexical knowledge, proving the effectiveness of our lexical knowledge transfer. Meanwhile, similar even more gain can also be observed in LED (w/ RT), once again proving that our method is complementary to distillation from a cross-encoder.
(3) In the queries that the LEX performs unfavorably (i.e., ground truth ranked lower than 100), LED and LED (w/ RT) are negatively impacted by the lexicon-aware teacher's mistakes. Interestingly, their original rankings of these ground truths are not very high, either. So these queries are intractable for both dense and lexicon-aware retrievers, which we leave for future work.

\begin{table}[t!]
\centering
\small
\caption{The average rank of golden passages by four retrievers on MS MARCO Dev. We bin the dev examples into buckets with the rank predicted by LEX and calculate the average ranking of other retrievers by group.
}
\begin{tabular}{lccccc}
\toprule
\multicolumn{1}{c}{\multirow{2}{*}{\textbf{Ranges}}} & \multicolumn{1}{c}{\multirow{2}{*}{\textbf{Count}}} & \multicolumn{4}{c}{\textbf{Average Ranking}}                                                                                                             \\ \cmidrule{3-6} 
\multicolumn{1}{c}{}                                 & \multicolumn{1}{c}{}                                & \multicolumn{1}{c}{\textbf{LEX}} & \multicolumn{1}{c}{\textbf{DEN}} & \multicolumn{1}{c}{\textbf{LED}} & \multicolumn{1}{c}{\textbf{LED (w/ RT)}} \\ \midrule
Top 1                                                & 1,787                                               & 1.0                                  & 2.5                                & 2.3                              & 2.4                                       \\
(1, 5]                                               & 2,242                                               & 3.1                                  & 6.4                                & 5.2                              & 4.8                                       \\
(5, 10]                                              & 875                                                 & 7.8                                  & 14.7                               & 13.8                             & 12.7                                      \\
(10, 50]                                             & 1,428                                               & 23.3                                 & 31.4                               & 31.1                             & 28.9                                      \\
(50, 100]                                            & 358                                                 & 70.5                                 & 80.0                               & 75.9                             & 74.0                                      \\
(100, 500]                                           & 445                                                 & 216.8                                & 156.3                              & 166.7                            & 154.1                                     \\
(500, 1000]                                          & 69                                                  & 698.0                                & 298.9                              & 334.7                            & 289.1                                     \\ \bottomrule
\end{tabular}

\label{tab:avg-rank}
\end{table}

\subsection{Case Study}
\begin{table}[!t]
\caption{Case study on MS-Marco Dev. `Passage+' denotes the golden passage of the corresponding query. `Rank' indicates the ranking of golden passage by retrievers.}
\vspace{2mm}
\resizebox{\linewidth}{!}{
\begin{tabular}{p{0.07\textwidth} | p{0.44\textwidth}}
\toprule
\textbf{Query}     & ID: 1090413// state the benefits of internet                                                                                                                                                                                                                                                                                                                                                                                                                                                                                                                                                                                                                                                                                                      \\ \hline
\textbf{Passage+}  & ID: 7998365// What Are Some Benefits of Using the Internet?<sep>Some of the \textbf{benefits of the Internet} include reduced geographical distance and fast communication. The Internet is also a hub of information where users can simply upload, download and publish ideas... \\ \hline
\textbf{Rank}      & LEX: 1; DEN: 3; LED: 1; LED w/ RT: 1                                                                                                                                                                                                                                                                                                                                                                                                                                                                                                                                                                                                                                                                                                                     \\ \hline
\textbf{Retrieved} & \textbf{DEN's 1st.} ID: 7339157 // -<sep>Advantages of the Internet. The Internet provides opportunities galore, and can be used for a variety of things. Some of the things that you can do via the Internet are: 1  E-mail: E-mail is an online correspondence system. 2  With e-mail you can send and receive instant electronic messages, which works like writing letters.                                                                                                                                                                                                                                                                                                                                                                                                   \\ \hline \hline

\textbf{Query}     & ID:1033652// what is the purpose of pencil tool                                                                                                                                                                                                                                                                                                                                                                                                                                                                                                                                                                                                                                                                                                            \\ \hline
\textbf{Passage+}  & ID: 7212314// Pencil<sep>Should I remove Pencil by Evolus Co? Pencil is built for the purpose of providing a free and \textbf{open-source GUI prototyping tool} that people can easily install and use to create mockups in popular desktop platforms.                                                                                                                                                                                                                                                                                                                                                                                                 \\ \hline
\textbf{Rank}      & LEX: 1; DEN: 11; LED: 1; LED w/ RT: 1                                                                                                                                                                                                                                                                                                                                                                                                                                                                                                                                                                                                                                                                                                                    \\ \hline
\textbf{Retrieved} & \textbf{DEN's 1st.} ID:313304 // Pencil<sep>This article is about the writing implement. For other uses, see Pencil (disambiguation). A pencil is a writing implement or art medium constructed of a narrow, solid pigment core inside a protective casing which prevents the core from being broken or leaving marks on the users hand during use.                      \\ \bottomrule
\end{tabular}
}
\label{tab:appendix-case-study}
\end{table}

Tab.~\ref{tab:appendix-case-study} shows the three case queries with rankings of 4 retrievers. With lexicon-aware capability, LED and LED w/ RT could retrieve golden passages as the top-1 result like their teacher LEX.
In particular, in the first case, DEN mismatches the ``\textit{benefits}'' in the query with ``\textit{advantages}'' in the passage since they are both positive words. On the contrary, the LEX and LED-series retrievers could exactly match the phrase ``\textit{benefits of the Internet}''.
In the second query, the ``\textit{Pencil tool}'' refers to a specific GUI prototyping tool (as highlighted in the positive passage). DEN misunderstands the mention ``\textit{Pencil tool}'' in the query and returns passages about the vanilla pencil, which is non-relevant to the user's intention.
The above two cases show that retrievers with lexicon-aware capability (i.e., LEX, LED, LED w/ RT) could well capture the salient phrases and entity mentions, providing more precise retrieval results.

\section{Conclusion}
In this paper, we consider developing a lexicon-enlightened dense retriever by transferring knowledge from a lexicon-aware sparse representation model into a dense one. To achieve this end, we propose to enlighten a dense retriever from two aspects, namely the lexicon-augmented contrastive objective and the pair-wise rank-consistent regularization.
Experimental results on three real-world retrieval benchmarks show that with a performance-comparable lexicon-aware representation model as the teacher, our strategies can improve a dense retriever consistently and significantly, even outdoing its teacher.
Further extensive analysis and discussions demonstrate the effectiveness and compatibility of our training strategies, as well as the interpretability of the LED retriever.




\bibliographystyle{ACM-Reference-Format}
\bibliography{anthology,custom}


\appendix
\clearpage
\section{Baselines}
\label{appendix-sec:baselines}
We compare with previous state-of-the-art baselines including traditional term-based techniques like BM25~\cite{Robertson2009BM25}, and dense as well as lexicon-aware retrievers. 
For lexicon-aware retrievers, DeepCT~\cite{Dai2019DeepCT} was trained to predict term weights.COIL~\cite{Gao2021COIL} used contextualized representation for exact term matching and UniCOIL~\cite{Lin2021UniCOIL} compressed vectors in COIL into scalars. DistilSPLADE-max and SPLADE-max~\cite{Formal2021SPLADEv2} were both trained with Eq.~\ref{eq:splade} and the latter one was further enhanced by a cross-encoder. The UniCOIL $\Lambda$ was a lexicon-aware model trained with UniCOIL's top-ranked passages and nagatives~\cite{Chen2021ImitateSparse}. For dense retrievers, ANCE~\cite{Xiong2021ANCE} selected hard training negatives from the entire collection.  ADORE~\cite{Zhan2021STAR-ADORE} used self-mining static negatives and then dynamic negatives.  TAS-B~\cite{Hofstatter2021TAS-B} proposed balanced topic-aware negative sampling strategies for effective teaching. CL-DRD~\cite{Zeng2022CL-DRD} taught the retriever in a curriculum learning fashion, starting from coarse-grained pair examples and progressing to fine-grained ones. ColBERTv1~\cite{Khattab2020ColBERT} and ColBERTv2~\cite{Khattab2021ColBERTv2} utilized late-interaction and the latter one further incorporates ranker distillation. TCT-ColBERT~\cite{Lin2021inbatch} utilized ColBERTv1 as the tightly-coupled teacher to enable in-batch distillation. The coCondenser~\cite{Gao2021coCondenser} augmented MLM loss with contrastive learning and based a model architecture~\cite{Gao2021Condenser} with a decoupled sentence and token interaction. PAIR~\cite{Ren2021PAIR} introduced passage-centric loss to assist the contrastive loss and combine cross-encoder teaching. RocketQAv2~\cite{Ren2021RocketQAv2} utilized K-L divergence to align the list-wise distributions between retriever and ranker and proposed hybrid data augmentation. AR2-G~\cite{Zhang2021AR2} used an adversarial framework to train the retriever and ranker simultaneously. Notably, AR2 used a different Recall@N evaluation from the official TREC Recall@N.\footnote{https://github.com/microsoft/AR2/tree/main/AR2} Therefore, we don't report their Recall@N performances in Tab.~\ref{tab:main-table}.

\clearpage

\end{document}